\pdfoutput=1

\documentclass[11pt]{article}

\usepackage[final]{acl}

\usepackage{times}
\usepackage{latexsym}

\usepackage[T1]{fontenc}

\usepackage[utf8]{inputenc}

\usepackage{microtype}
\usepackage{microtype}
\usepackage{amsfonts}
\usepackage{amsmath}
\usepackage{enumitem}
\usepackage{graphicx}
\usepackage{float}
\usepackage{booktabs}
\usepackage{subcaption} 
\usepackage{colortbl}
\usepackage{xcolor}

\definecolor{lightcolor}{gray}{0.85} 
\definecolor{darkcolor}{gray}{0.75} 

\usepackage{inconsolata}

\title{EventRL: Enhancing Event Extraction with Outcome Supervision for Large Language Models}

\author{ Jun Gao\textsuperscript{1}\enskip Huan Zhao\textsuperscript{2} \enskip Wei Wang\textsuperscript{3}\enskip Changlong Yu\textsuperscript{4}\enskip Ruifeng Xu\textsuperscript{1}\\
\normalsize \textsuperscript{1}Harbin Institute of Technology (Shenzhen) \\
\normalsize \texttt{imgaojun@gmail.com}\quad \texttt{xuruifeng@hit.edu.cn}\\
\normalsize   \textsuperscript{2}4Paradigm. Inc. \quad \textsuperscript{3}Tsinghua University \quad
\textsuperscript{4}HKUST, Hong Kong, China\\
\normalsize \texttt{zhaohuan@4paradigm.com}  \quad \texttt{weiwangorg@163.com} \quad \texttt{cyuaq@cse.ust.hk}\\
}

\begin{document}
\maketitle
\begin{abstract}

In this study, we present EventRL, a reinforcement learning approach developed to enhance event extraction for large language models (LLMs). EventRL utilizes outcome supervision with specific reward functions to tackle prevalent challenges in LLMs, such as instruction following and hallucination, manifested as the mismatch of event structure and the generation of undefined event types. We evaluate EventRL against existing methods like Few-Shot Prompting (FSP) (based on GPT4) and Supervised Fine-Tuning (SFT) across various LLMs, including GPT-4, LLaMa, and CodeLLaMa models. Our findings show that EventRL significantly outperforms these conventional approaches by improving the performance in identifying and structuring events, particularly in handling novel event types. The study emphasizes the critical role of reward function selection and demonstrates the benefits of incorporating code data for better event extraction. While increasing model size leads to higher accuracy, maintaining the ability to generalize is essential to avoid overfitting. 

\end{abstract}

\section{Introduction}
Event extraction, a crucial task in natural language processing (NLP), aims at identifying and categorizing events within texts~\citep{chen2015event,nguyen2016joint,liu2018jointly, yang2019exploring,lu2021text2event,gao2023exploring}. 
\begin{figure}[ht]
    \centering
    \includegraphics[width=0.45\textwidth]{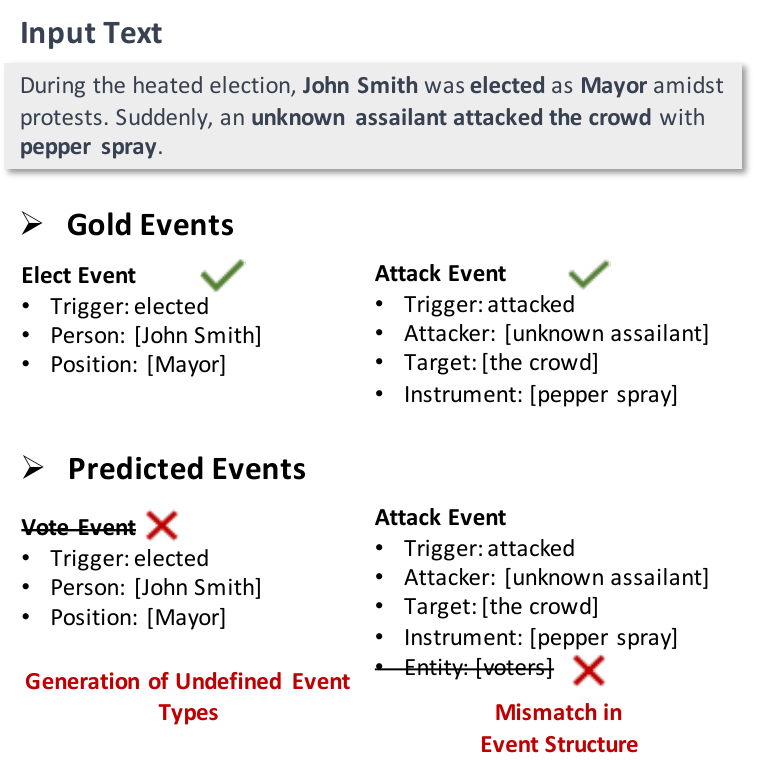}
    \caption{Examples of common errors in LLM-Based event extraction. The left side depicts an error of generating an undefined event type, specifically an unexpected ``Vote'' event not included in the guidelines. The right side shows a structural mismatch error within an ``Attack'' event, incorporating an "Entity" argument that deviates from the pre-defined event schema.}
    \label{fig:error_cases}
    \vspace{-1.5em}

\end{figure}
Recently, large language models~(LLMs) have demonstrated impressive capabilities in language understanding and generation for various tasks~\citep{ouyang2022training,chen2021evaluating,achiam2023gpt,zhao2023survey}.
However, they also encounter specific challenges, such as instruction following~\cite{zhou2023instruction,zeng2023llmbar} and the generation of inaccurate or irrelevant content, often referred to as hallucinations~\cite{li2023halueval}.

In the context of event extraction tasks, these models encounter similar difficulties, including mismatches in event structure and the generation of undefined event types~\citep{gao2023exploring}. 
As illustrated in Figure~\ref{fig:error_cases},
a mismatch in event structure refers to inaccuracies like incorrectly including an irrelevant argument. For instance, an ``Attack'' event erroneously includes a non-existent ``entity'' role. The generation of undefined event types refers to the model's prediction of event types that are not predefined in the task instruction. For example, in texts related to elections, the model might unpredictably predict a ``Vote'' event type, which is not defined in the task instruction.
These issues can be seen as manifestations of instruction following and hallucination problems within the realm of event extraction. Recent studies~\citep{wang2023instructuie,gao2023benchmarking,sainz2023gollie} have attempted to address these challenges using Supervised Fine-Tuning (SFT) methods, but the performance has been far from satisfactory. A potential reason for this is that event extraction demands too much recognition for abstract concepts and relations, and it suggests that there is a need to focus more on enhancing the high-level understanding for event comprehension~\citep{huang2023reevaluation}.

A major limitation of SFT approaches in event extraction is their inability to accurately recognize errors in event structures, such as incorrect argument inclusion or predicting events not defined in the guidelines. 
This issue may stem from the reliance on Negative Log Likelihood (NLL) loss, which, while effective for general language modeling, falls short in capturing the intricacies of event extraction. Specifically, when it comes to event extraction, both types of errors—incorrect predictions of event types and incorrect predictions of an event argument's role—often differ from the correct samples by just a single word in the text. However, in terms of NLL loss, these errors result in only minor differences, failing to reflect their significant impact on event extraction performance. This discrepancy is particularly critical because an error in predicting the event type can lead to a cascade of errors in the extraction of all associated arguments, drastically reducing the accuracy of the entire extraction process. Therefore, while NLL loss might marginally penalize these mistakes, their actual consequences are far more severe.

One potential solution is to integrate feedback on the model's performance in identifying and structuring events into its training process. 
This method, known as outcome supervision, draws inspiration from previous works in solving math problems~\citep{uesato2022solving,lightman2023let}.
By incorporating outcome-based feedback, the model can adjust and refine its strategies for more accurate event identification and structuring, addressing the specific challenges it faces in understanding and extracting events from text.
To this end, we introduce EventRL, a novel reinforcement learning framework designed to enhance event extraction by directly responding to the accuracy of the model's output. EventRL addresses key issues faced by LLMs in event extraction, such as the mismatch in event structures and the generation of undefined event types. It leverages outcome performance as feedback to penalize errors, guiding the model to adjust its strategies for better performance.
We explore three event-specific reward functions: Argument-F1, Average-F1, and Product-F1. These functions are designed to improve the model's comprehension of event structures
 Moreover, to enhance training stability, we introduce Teacher-Force Threshold and Advantage Clipping strategies, mitigating policy degradation and preventing catastrophic forgetting.
Our contributions can be summarized as follows:
\begin{itemize}[leftmargin=*]
    \item We introduce outcome supervision to LLMs for event extraction, focusing on task outcomes to improve event understanding and extraction. To the best of our knowledge, we are the first to incorporate outcome feedback on event structures into the training process of LLMs for EE. 
    \item We develop EventRL, a novel approach that implements outcome supervision through reinforcement learning with tailored reward functions, to provide a more precise and targeted training method for EE.
    \item Extensive experiments with LLMs of varying sizes show that EventRL significantly outperforms standard SFT methods. Notably, EventRL shows remarkable improvements in handling unseen events and significantly reduces errors in event structure and type definition, thereby validating the effectiveness of outcome supervision in boosting the capabilities of LLMs in EE.
\end{itemize}

\begin{figure*}[ht]
    \centering
    \includegraphics[width=0.9\textwidth]{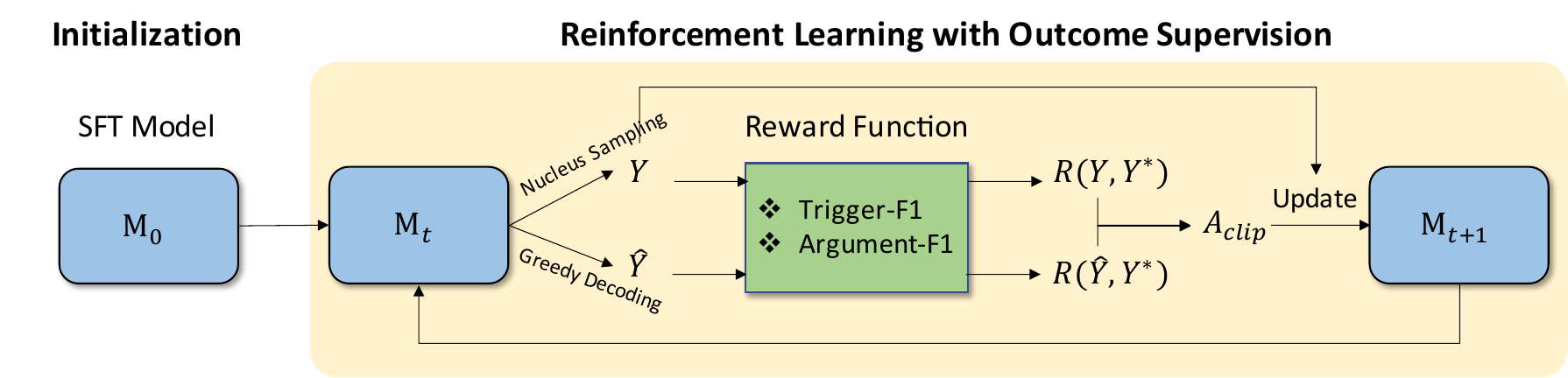}
    \caption{The EventRL framework architecture, demonstrating the process from initialization with an SFT Model \( M_0 \), through iterative model updates \( M_t \) to \( M_{t+1} \) via Outcome Supervision. This includes using reinforcement learning with reward functions based on Trigger-F1 and Argument-F1 scores, which guide policy gradient updates for enhanced event extraction from text.}
    \label{fig:arch}
    \vspace{-1em}

\end{figure*}


\section{Related Work}

\paragraph{Event Extraction}
Event Extraction (EE) has evolved from traditional sequence labeling methods to the integration of advanced machine learning models, particularly large language models. The initial approach to EE focused on word-level classification, capturing sentence dependencies~\citep{chen2015event,nguyen2016joint,liu2018jointly, yang2019exploring,wadden2019entity}. A significant shift occurred with the introduction of Machine Reading Comprehension techniques, which transformed EE into a question-answering task, enhancing event extraction~\citep{chen2019reading,du2020event,li2020event,zhou2021role,wei2021trigger}. The subsequent development of sequence-to-structure generation with Transformer-based architectures further streamlined the process by merging event detection and argument extraction~\citep{lu2021text2event, lu2022unified,lou2023universal}. The latest advancement involves LLMs, which, due to their extensive pre-training, demonstrate exceptional generalization capabilities, advancing traditional EE techniques and enabling zero-shot event extraction, marking a notable progression in NLP~\citep{wei2023zero,gao2023exploring,wang2023instructuie, sainz2023gollie, gao2023benchmarking}.
In contrast to previous work, our work focuses on utilizing outcome supervision to refine model training of LLMs for event extraction, thereby enhancing performance. Notably, we are the first to incorporate outcome feedback into the LLM training process for EE.

\paragraph{Large Language Models and Outcome Supervision}
Large language models has marked a significant advance in the field of NLP.
Recent studies have demonstrated the exceptional capability of LLMs, such as ChatGPT and GPT-4, to perform with remarkable performance in event extraction~\citep{gao2023exploring,sainz2023gollie}. These models demonstrate notable performance gain even in zero-shot learning settings, indicating their potential to generalize across different types of event-related information without the need for task-specific training data.
Despite the progress, LLMs continue to face challenges related to instruction following~\citep{ouyang2022training} and hallucinations~\citep{huang2023survey,zhang2023siren,li2024dawn}. To address these issues, researchers have explored a range of strategies, including Supervised Fine-Tuning (SFT)~\citep{wang2022super,zhang2023instruction,wang2023multitask}, Reinforcement Learning from Human Feedback~(RLHF)~\citep{stiennon2020learning,ouyang2022training,kaufmann2023survey}, and more recently, approaches that combine outcome supervision with reinforcement learning techniques~\citep{uesato2022solving, lightman2023let, yu2023outcome} for solving math problems. 
Building on these seminal works, our work makes the first attempt to introduce outcome supervision for event extraction tasks, which can harness the power of LLMs while directly address their limitations in instruction following and hallucinations, thus significantly improving the efficacy and reliability of event extraction.

\section{EventRL}
\subsection{Overview}
EventRL aims to enhance event extraction through outcome supervision with reinforcement learning for LLMs. As depicted in Figure~\ref{fig:arch}, EventRL begins with an SFT phase to establish a baseline understanding of event extraction. It then progresses to implement outcome supervision, leveraging reward functions based on resulting performance~(Trigger-F1 and Argument-F1 scores) to guide the model's training via reinforcement learning techniques. 
To ensure stable and effective learning, EventRL incorporates stabilization strategies, including a Teacher-Force Threshold and Advantage Clipping, which are critical for mitigating policy degradation and preventing catastrophic forgetting. 
The subsequent sections will delve into the detailed implementation of these components.

\subsection{Initialization}
\begin{figure}[ht]
    \centering
    \includegraphics[width=0.45\textwidth]{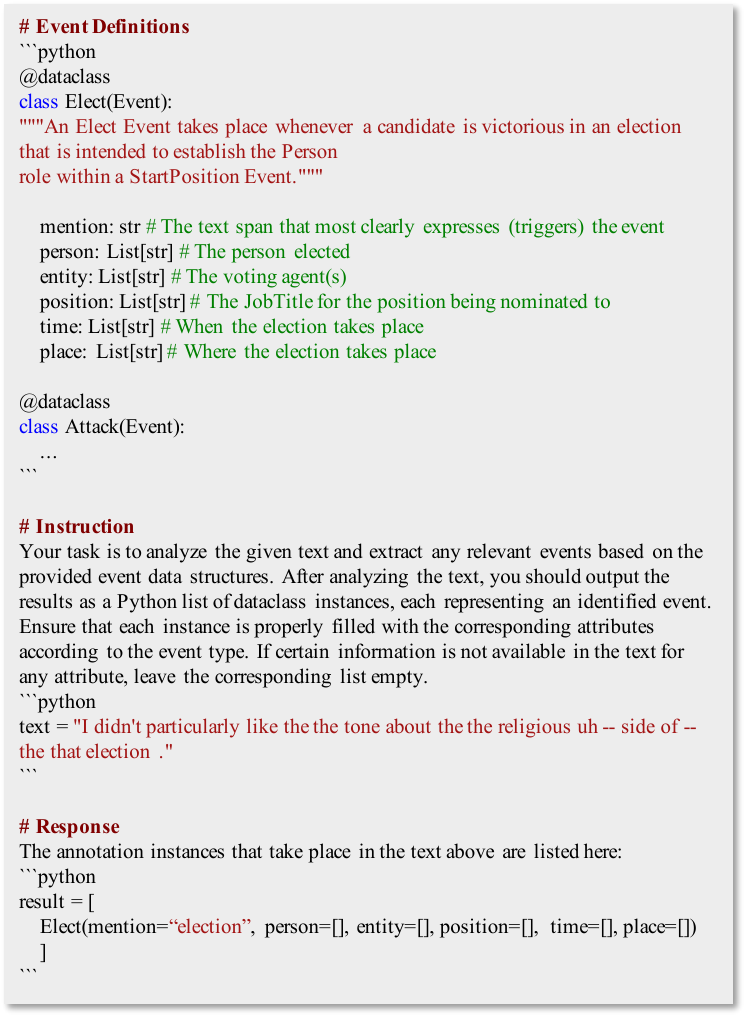}
    \caption{Illustration of the input-output format in EventRL for event extraction. The input includes Event Definitions in Python dataclass format and a natural language instruction. The output showcases a Python list of dataclass instances as the Response, representing extracted events from the given text. The complete event definitions can be found in Figure~\ref{fig:definitions} in Appendix.}
    \label{fig:io_example}
    \vspace{-1em}
\end{figure}

\paragraph{Input and Output Format}
In the Initialization phase of EventRL, we adopt a hybrid input and output format, building upon the work of \citet{sainz2023gollie}. As illustrated in Figure~\ref{fig:io_example}, our approach combines structured Python dataclass formats for event definitions with natural language instructions for task descriptions. This allows for precise event representation while maintaining user-friendly instructions. The output is a Python list of dataclass instances, representing the extracted events in a structured and programmatically accessible format. This hybrid format enhances the model's ability to process and output complex event information accurately, bridging the gap between structured coding and natural language understanding.

\paragraph{Supervised Fine-tuning}

The SFT process~\citep{wang2022super,zhang2023instruction,wang2023multitask,peng2023instruction} in our approach is an important initial phase that establishes a foundational understanding of the event extraction process in the model. During SFT, the model is trained using a labeled dataset that consists of examples, where each event is explicitly defined with its corresponding triggers and arguments. This dataset provides clear and structured examples of desired event extraction outputs.

\subsection{Outcome Supervision with RL}
\paragraph{Problem Formulation}
To implement outcome supervision, we leverage reinforcement learning~\citep{kaelbling1996reinforcement}. 
The RL method treats event extraction as a sequential decision-making process, where the model \(M\), guided by its policy \(\pi_{\theta}\), generates predictions \(Y\), which include event triggers and their arguments, based on inputs \(X\). 

The model's parameters are updated to maximize expected rewards, leveraging an advantage-based policy optimization method to guide learning~\citep{sutton2018reinforcement}. The update rule is as follows: 
\begin{equation}
    \Delta\theta \propto \nabla_{\theta} \log \pi_{\theta}(Y|X) \cdot A,
\end{equation}
where \(\nabla_{\theta} \log \pi_{\theta}(Y|X)\) is the gradient of the log-probability of taking action \(Y\) in state \(X\) under policy \(\pi_{\theta}\). The advantage function is \(A = R(Y, Y^*) - b\), where \(R(Y, Y^*)\) is the reward function and \(b\) is a baseline reward. This function calculates the difference between the reward \(R(Y, Y^*)\) for the model's predictions and a baseline \(b\), identifying actions that yield above-average benefits.
This approach stabilizes policy gradient estimates by reducing variance~\citep{rennie2017self}.

\paragraph{Reward Function}
In our EventRL framework, the reward function focuses on two primary aspects: Trigger Extraction and Argument Extraction. These aspects are quantified through Trigger-F1 and Argument-F1 scores, respectively, which serve as the basis for our reward function. The Trigger-F1 score assesses the model's ability to accurately identify and classify event triggers, while the Argument-F1 score evaluates the precision in identifying and classifying the arguments associated with those triggers.
Following previous work~\citep{lu2021text2event}, we adopt the following criteria of the evaluation:
A trigger is correct if its event type matches with the ground truth. Similarly, an argument is considered correctly identified if its event type and role match with the ground truth. 

In our work, we explore three different reward function designs, shown in the following equation:
\begin{equation}
R = 
\begin{cases} 
F1_{\text{arg}}, & \text{Argument-F1} \\
\frac{1}{2}(F1_{\text{trig}} + F1_{\text{arg}}), & \text{Average-F1} \\
F1_{\text{trig}} \times F1_{\text{arg}}, & \text{Product-F1}
\end{cases}
\end{equation}
The Argument-F1 reward, \(F1_{\text{arg}}\), focuses on how well the model can identify and classify the arguments of events. Meanwhile, the Average-F1 reward, \(\frac{1}{2}(F1_{\text{trig}} + F1_{\text{arg}})\), gives a balanced look by combining Trigger-F1 and Argument-F1 scores, offering a full view of the effectiveness of the model. Lastly, the Product-F1 reward, \(F1_{\text{trig}} \times F1_{\text{arg}}\), highlights the importance of doing well in both trigger detection and argument extraction by multiplying these scores together, pushing the model to excel in both areas for a better total reward.

In current approaches of instruction tuning, especially in response generation tasks, feedback often comes from model scoring. This can involve predictions from a reward model trained on human preference data~\citep{ouyang2022training} or direct scoring by more advanced models like GPT-4~\citep{cui2023ultrafeedback}. Although, for event extraction tasks, we could train a reward model on positive and negative examples or have GPT-4 score the outputs, the necessity for such methods is reduced. This is because standard evaluation metrics for event extraction already provide a clear reflection of output quality. However, there is still value in the model generating natural language judgements~\citep{xu2023reasons} to further address and correct training issues. We leave this for future work.

\paragraph{Advantage Calculation}
To calculate the advantage, we compare the rewards of two distinct strategies for extracting events: greedy decoding and nucleus sampling. 
When processing a given text, the model generates two outputs: one through greedy decoding (\(\hat{Y}\)) and another via nucleus sampling (\(Y\)). We then assess these outputs by calculating their rewards, \(R(\hat{Y},Y^*)\) for the greedy decoding output and \(R(Y,Y^*)\) for the nucleus sampling output. The advantage function \(A(\cdot)\) quantifies the benefit of the nucleus sampling strategy over the greedy decoding by measuring the difference in their rewards:
\begin{equation}
    A = R(Y,Y^*) - R(\hat{Y},Y^*).
\end{equation}
This comparison highlights the effectiveness of exploratory actions in improving event extraction outcomes compared to the baseline approach.

\subsection{Stabilization Strategies in EventRL}
To ensure stable training, two key stabilization strategies are implemented: the Teacher-Force Threshold~\citep{bengio2015scheduled} and Advantage Clipping~\citep{schulman2017proximal}.
\paragraph{Teacher-Force Threshold}
The Teacher-Force Threshold strategy is employed to mitigate policy degradation, especially in instances where the model's performance on certain samples is significantly below par. This strategy involves setting a threshold value, denoted as \(\tau\), for the model's performance score (typically measured using the outcome score from greedy decoding). When the model's performance on a given sample falls below \(\tau\), the learning process is adjusted to ``teacher forcing'' mode. In this mode, the model is temporarily guided using the gold events $Y^*$ instead of its own generated output $Y$, effectively providing a more reliable learning signal. 

\paragraph{Advantage Clipping}
Advantage Clipping in the EventRL is specifically designed to address the challenge of catastrophic forgetting, a phenomenon where the model's performance on previously learned tasks deteriorates as it focuses on new ones. This issue often arises when the advantage values for certain samples are too low, leading to negligible updates and causing the model to overlook these samples during training.
The strategy involves setting a lower bound for the advantage values, denoted as \(A_{\text{min}}\). This lower bound ensures that every sample, regardless of its initial advantage value, contributes a minimum threshold of influence to the learning process. 
The advantage clipping process is reformulated to focus solely on the lower bound, as follows: $A_{\text{clip}} = \text{max}(A, A_{\text{min}})$.

\begin{table*}[ht]
\centering
\small
\label{tab:performance_comparison}
\begin{tabular}{lcccccc}
\toprule
 & \multicolumn{3}{c}{\textbf{Held-in test}} & \multicolumn{3}{c}{\textbf{Held-out test}}\\
\cmidrule(r){2-4} \cmidrule(lr){5-7}
\textbf{Method}& \textbf{Trigger} & \textbf{Argument} & \textbf{AVG} & \textbf{Trigger} & \textbf{Argument} & \textbf{AVG} \\
\midrule
GPT4 + FSP~(0-Shot) & 6.04 & 22.08 & 14.06 & 15.42 & 17.69 & 16.56  \\
GPT4 + FSP~(1-Shot) & 23.02 & 22.82 & 22.92 & 19.32 & 17.83 & 18.58  \\
GPT4 + FSP~(2-Shot) & 24.65 & 23.48 & 24.06 & 24.12 & 18.31 & 21.22  \\
GPT4 + FSP~(3-Shot) & 31.58 & 23.53 & 27.55 & 27.07 & 18.61 & 22.84  \\
\midrule
LLaMa-7B + SFT & 71.33 & 40.74 & 56.03 & 48.51 & 26.18 & 37.35  \\
LLaMa-7B + EventRL (Arg-F1) & \cellcolor{darkcolor}73.06 & \cellcolor{lightcolor}42.34 & \cellcolor{lightcolor}57.70 & 51.15 & 29.32 & 40.23  \\
LLaMa-7B + EventRL (AVG-F1) & \cellcolor{lightcolor}72.34 & 42.29 & 57.32 & \cellcolor{darkcolor}54.59 & \cellcolor{lightcolor}29.81 & \cellcolor{darkcolor}42.20  \\
LLaMa-7B + EventRL (Prod-F1) & 72.03 & \cellcolor{darkcolor}49.41 & \cellcolor{darkcolor}60.72 & \cellcolor{lightcolor}51.71 & \cellcolor{darkcolor}29.97 & \cellcolor{lightcolor}40.84  \\
\midrule
LLaMa-13B + SFT & 76.23 & 51.16 & 63.69 & 51.61 & 32.46 & 42.04  \\
LLaMa-13B + EventRL (Arg-F1) & \cellcolor{darkcolor}77.61 & \cellcolor{lightcolor}51.93 & \cellcolor{lightcolor}64.77 & \cellcolor{lightcolor}53.07 & 32.83 & 42.95  \\
LLaMa-13B + EventRL (AVG-F1) & \cellcolor{lightcolor}77.26 & \cellcolor{darkcolor}\cellcolor{darkcolor}54.55 & \cellcolor{darkcolor}65.90 & 51.53 & \cellcolor{lightcolor}34.93 & \cellcolor{lightcolor}43.23  \\
LLaMa-13B + EventRL (Prod-F1) & 76.23 & 51.66 & 63.94 & \cellcolor{darkcolor}53.79 & \cellcolor{darkcolor}35.03 & \cellcolor{darkcolor}44.41  \\
\midrule
CodeLLaMa-7B + SFT & 74.31 & 44.16 & 59.23 & \cellcolor{darkcolor}62.21 & 37.26 & 49.74  \\
CodeLLaMa-7B + EventRL (Arg-F1) & 75.35 & \cellcolor{darkcolor}50.84 & \cellcolor{darkcolor}63.09 & \cellcolor{lightcolor}61.64 & 37.93 & 49.78  \\
CodeLLaMa-7B + EventRL (AVG-F1) & \cellcolor{darkcolor}77.14 & 47.06 & 62.10 & 61.62 & \cellcolor{lightcolor}39.01 & \cellcolor{darkcolor}50.32  \\
CodeLLaMa-7B + EventRL (Prod-F1) & \cellcolor{lightcolor}76.60 & \cellcolor{lightcolor}48.39 & \cellcolor{lightcolor}62.49 & 60.34 & \cellcolor{darkcolor}39.69 & \cellcolor{lightcolor}50.01  \\
\midrule
CodeLLaMa-13B + SFT & \cellcolor{lightcolor}77.70 & 47.21 & 62.46 & 61.40 & 41.98 & 51.69  \\
CodeLLaMa-13B + EventRL (ARG-F1) & \cellcolor{darkcolor}80.88 & \cellcolor{lightcolor}50.51 & \cellcolor{darkcolor}65.69 & \cellcolor{lightcolor}62.56 & 41.39 & \cellcolor{lightcolor}51.97  \\
CodeLLaMa-13B + EventRL (AVG-F1) & 76.98 & 48.18 & 62.58 & 60.86 & \cellcolor{darkcolor}42.37 &  51.62 \\
CodeLLaMa-13B + EventRL (Prod-F1) & 77.03 & \cellcolor{darkcolor}50.78 & \cellcolor{lightcolor}63.91 & \cellcolor{darkcolor}62.57 & \cellcolor{lightcolor}42.14 & \cellcolor{darkcolor}52.35  \\
\midrule
CodeLLaMa-34B + SFT & 74.65 & 56.69 & 65.67 & 57.98 & 39.52 & 48.75 \\ \bottomrule
\end{tabular}
\caption{Performance comparison of different Large Language Models (LLMs), including LLaMa, CodeLLaMa, and GPT4 on the ACE05 dataset. We highlight the best performance in dark gray and the second-best in light gray. We evaluate FSP method with GPT4 using 0, 1, 2, 3 example settings, where examples are randomly sampled from the training set. For FSP (0-Shot), both Held-in and Held-out tests involve unseen event types, while for other FSP methods with examples, Held-in tests use seen event types, and Held-out tests use unseen event types. Additionally, we compare the performance of CodeLLaMa-34B with SFT, noting that due to limited computational resources, the EventRL method was not applied to CodeLLaMa-34B.}
\label{tab:main_results}
\vspace{-1em}
\end{table*}

\section{Experimental Setup}
\subsection{Dataset and Data Splitting Strategy}
To evaluate our model's performance comprehensively, we conducted experiments on the ACE05 dataset~\citep{ldcace05}, known for its diversity in event types. This allows us to test the model's capability in extracting both familiar (seen) and novel (unseen) event types effectively. The ACE05 dataset, which contains a total of 33 event types, was used to construct our experimental setup. We selected 7 event types for the training set, validation set, and the held-in test set (seen event types). We then chose 19 different event types to form the held-out test set (unseen event types), ensuring a rigorous evaluation of the model's generalization abilities. To maintain a balanced dataset, we sampled 50 instances for each event type in the training set, 10 for each in the validation set, and 20 for each in both the held-in and held-out test sets. This strategy ensures that the model is trained and evaluated under varied conditions, providing a comprehensive understanding of its performance across different event types. For detailed statistical information, please refer to Appendix~\ref{app:dataset}.

\subsection{Comparison Methods}
In our study, we assess the efficacy of EventRL against current methods like Few-Shot Prompting (FSP) and Supervised Fine-Tuning (SFT), applying these methods across various LLMs to evaluate performance in event extraction:
\textbf{(1) Few-Shot Prompting (FSP):} 
Implemented on the specific version of \textbf{GPT-4} (API version 2023-05-15), provided by the Azure OpenAI Service, this method relies on the model's intrinsic capabilities by providing a limited set of examples before task execution. 
\textbf{(2) Supervised Fine-Tuning (SFT):} 
This approach involves direct training of models on specific datasets, with feedback provided via NLL loss. SFT experiments were primarily conducted on \textbf{LLaMa} variants (LLaMa2-7B and LLaMa2-13B) and \textbf{CodeLLaMa} models (7B, 13B, and 34B versions).
\textbf{(3) EventRL with Proposed Reward Functions:}
Our proposed EventRL framework was evaluated in three distinct configurations, each utilizing a different reward function designed to optimize the model's performance in event extraction: \textbf{EventRL (Arg-F1)} utilizes Argument-F1 as feedback, \textbf{EventRL (Avg-F1)} aims to balance Trigger-F1 and Argument-F1, and \textbf{EventRL (Prod-F1)} seeks to maximize the product of Trigger-F1 and Argument-F1. These variants were tested across \textbf{LLaMa}~\citep{touvron2023llama} and \textbf{CodeLLaMa}~\citep{roziere2023code} models to investigate their effectiveness in event extraction.

Implementation details of different methods can be found in Appendix~\ref{app:implem_details}.

\section{Experimental Results}

\subsection{Main Results}

\paragraph{Overall Performance}
Table~\ref{tab:main_results} presents a comprehensive performance comparison of different LLMs, including GPT4, LLaMa, and CodeLLaMa variants, across various training approaches such as SFT (Supervised Fine-Tuning), FSP (Few-Shot Prompting), and our proposed EventRL methods. 
The table reports Trigger and Argument F1 scores, alongside their averages (denoted as AVG), which are calculated as the mean of the respective Trigger and Argument F1 scores for each method, both for events seen during training (Held-in test) and for novel, unseen event types (Held-out test).
Our findings show that EventRL outperforms both SFT and FSP methods in event extraction. Specifically, when comparing the AVG scores, EventRL demonstrates superior overall performance. For example, in the Held-in test, the EventRL (Prod-F1) method using the LLaMa-7B model achieves an AVG score of 60.72, surpassing the SFT method's 56.03. Similarly, in the Held-out test, EventRL (Prod-F1) with LLaMa-7B reaches an AVG score of 40.84, compared to 37.35 by SFT. This indicates EventRL's effectiveness in accurately identifying event structures and predicting events.

Moreover, EventRL shows remarkable generalization capabilities, especially in handling unseen event types. Using the LLaMa-13B model, EventRL (Prod-F1) scores an AVG of 44.41 in the Held-out test, outperforming the SFT method's 42.04. These results highlight EventRL's robustness and its ability to adapt to new, unseen event types better than the other methods.
The success of EventRL can be attributed to its specialized reward functions (Arg-F1, AVG-F1, and Prod-F1), which provide targeted feedback for refining the model's understanding and extraction of events. This tailored approach ensures that EventRL not only excels in extracting events from texts but also adapts effectively across different model sizes and architectures, including LLaMa and CodeLLaMa variants.

\paragraph{Impact of Different Reward Functions}
As shown in Table~\ref{tab:main_results}, the choice of reward functions significantly influences the performance of LLMs in event extraction, with AVG-F1 and Prod-F1 rewards demonstrating clear advantages. For the LLaMa-7B model, the Prod-F1 reward function yielded the best AVG Score, reaching 60.72 in Held-in test and 40.84 in Held-out test. This indicates that focusing on the interdependence of trigger and argument performance through the Prod-F1 reward enhances the model's overall ability to accurately extract events.
In the case of the larger LLaMa-13B model, the AVG-F1 reward function achieved an AVG Score of 65.90 in Held-in test, while the Prod-F1 function excelled in Held-out test with an AVG Score of 44.41. This performance trend is also reflected in the CodeLLaMa models, where the Prod-F1 reward again demonstrated its effectiveness, particularly achieving a 52.35 AVG Score in the Held-out test for the CodeLLaMa-13B model. These results underline the importance of carefully selecting reward functions to optimize the event extraction capabilities of LLMs, with Prod-F1 and AVG-F1 rewards proving to be particularly beneficial in fostering a deeper understanding and extraction of events from texts.

\subsection{Further Analysis}
\label{sec:further_analysis}

\paragraph{Ablation Study}
Table~\ref{tab:ablation_teacher_force} presents an ablation study focusing on key features of the EventRL approach, particularly looking at the impact of removing Teacher-Forcing and Advantage-Clipping on event extraction performance using the ACE05 dataset.
As can be seen, removing Teacher-Forcing led to a significant performance drop, with the Trigger-F1 score decreasing from 73.06 to 65.38 and the Argument F1 score from 42.34 to 27.25 for the Argument-F1. Similarly, excluding Advantage-Clipping resulted in a decline, notably in the Argument-F1 score, from 42.34 to 39.68. 
These results highlight the critical role of both strategies in ensuring EventRL's effectiveness and stability for event extraction.
More analysis of these two components can be found in Appendix~\ref{app:analysis_eventrl}.

\begin{table}[t]
\centering
\small
\begin{tabular}{lcccc}
\toprule
 & \multicolumn{2}{c}{\textbf{Held-in test}} & \multicolumn{2}{c}{\textbf{Held-out test}}\\
\cmidrule(r){2-3} \cmidrule(lr){4-5}
\textbf{Method}& \textbf{Trig.} & \textbf{Arg.} & \textbf{Trig.} & \textbf{Arg.} \\
\midrule
SFT&71.33&40.74&48.51&26.18 \\
\midrule
EventRL (Arg-F1)&73.06&42.34&51.15&29.32 \\
\hspace{3mm}w/o Tearcher-Force&65.38&27.25&47.34&19.28 \\
\hspace{3mm}w/o Advantage-Clip &67.60&39.68&46.37&24.21 \\
\midrule
EventRL (Prod-F1) &72.03&49.41&51.71&29.97 \\
\hspace{3mm}w/o Tearcher-Force &67.81&36.01&46.24&16.12 \\
\hspace{3mm}w/o Advantage-Clip &72.20&44.60&49.11&21.24 \\
\bottomrule
\end{tabular}
\caption{Ablation Study on ACE05 dataset using LLaMa-7B.}
\label{tab:ablation_teacher_force}
\end{table}

\begin{figure}[t]
    \centering
    \includegraphics[width=0.45\textwidth]{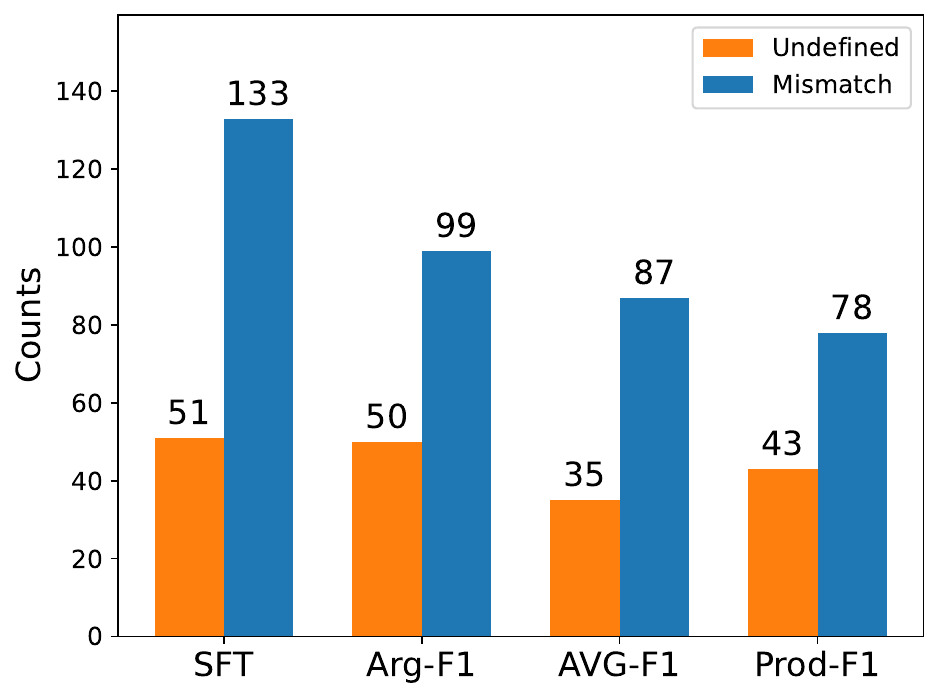}
    \caption{This chart quantifies the error counts for undefined event types and structural mismatches in event extraction on the LLaMa-7B model, comparing SFT with three EventRL training methods: Arg-F1, AVG-F1, and Prod-F1.}
    \label{fig:error_stats}
    \vspace{-1em}
\end{figure}

\paragraph{Error Analysis}
Figure~\ref{fig:error_stats} presents a comparative analysis of error types in event extraction when utilizing different training methods on the LLaMa-7B model. The Supervised Fine-Tuning (SFT) method resulted in a notably high occurrence of \textit{undefined event type} errors, totaling 133, and structural mismatch errors, at 51. In contrast, the EventRL (Arg-F1) reduced ``Undefined'' errors to 99 and marginally decreased ``Mismatch'' errors to 50. A more significant improvement is observed with the EventRL (AVG-F1) approach, which cut down ``Undefined'' errors to 87 and ``Mismatch'' errors to the lowest count of 35, indicating a superior balance in error mitigation. The EventRL (Prod-F1) also demonstrated improvement, lowering ``Undefined'' errors to 78 and ``Mismatch'' errors to 43, although not as effectively as AVG-F1. These numbers highlight the effectiveness of the EventRL training methods in reducing errors in large language model training for event extraction.

\paragraph{Code Data Pretraining Enhances Event Extraction Performance}
Table~\ref{tab:main_results} reveals that models enhanced with code data, notably CodeLLaMa, significantly outperform their counterparts without code enhancement, like LLaMa, in event extraction. Specifically, at the 7B scale, CodeLLaMa (SFT) achieved an AVG score of 59.23 in Held-in test and 49.74 in Held-out test, surpassing LLaMa's 56.03 and 37.35, respectively. This improvement illustrates the positive impact of coding capabilities on the model's ability to extract events, both seen and unseen.
When scaling up to 13B, the gap in performance between CodeLLaMa and LLaMa widens further, especially in the Held-out test, where CodeLLaMa (SFT) scores an AVG of 51.69, compared to LLaMa's 42.04. Additionally, employing the EventRL method with CodeLLaMa leads to superior outcomes across different reward function setups, demonstrating that code data enhancement not only boosts the model's understanding of structured information but also enhances its adaptability and accuracy in complex tasks like event extraction.

\paragraph{Analysis on  Model Scale}
 From Table~\ref{tab:main_results}, we observe a clear trend that increasing model scale positively impacts event extraction performance. For instance, when comparing the EventRL (Prod-F1) approach, the performance in terms of the AVG score improves significantly as we move from a 7B parameter model to a 13B parameter model, from 60.72 to 65.90 in the Held-in test. This indicates that larger models have enhanced capabilities in processing and understanding complex language structures, which leads to more accurate event extraction.
However, scaling the model size further to 34B parameters introduces the risk of overfitting, especially evident in the Held-out test. For example, the CodeLLaMa-34B model, under the SFT approach, shows an AVG score of 65.67 in Held-in test but drops to 48.75 in Held-out test, indicating a decline in the model's ability to generalize to unseen event types. 
\begin{figure}[]
    \centering
    \includegraphics[width=0.45\textwidth]{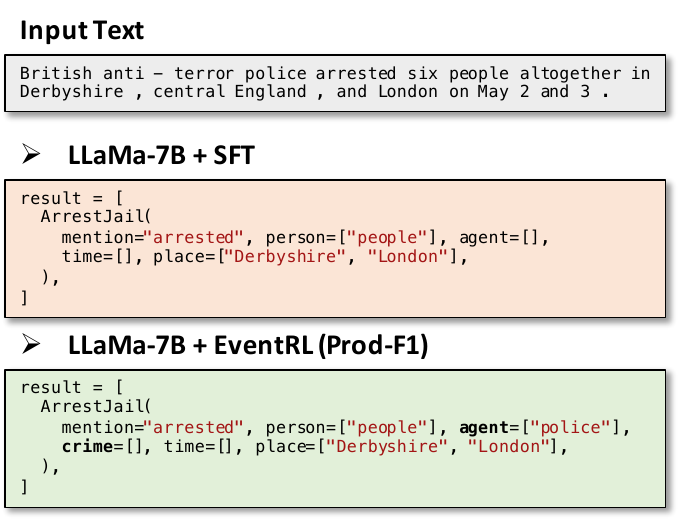}
    \caption{A comparison of event extraction results between LLaMa-7B + SFT and LLaMa-7B + EventRL (Prod-F1). Note that here the results of EventRL (Prod-F1) are totally accurate.}
    \label{fig:case_study}
    \vspace{-1em}
\end{figure}

\paragraph{Case Study}
Figure~\ref{fig:case_study} showcases a comparison between the results of LLaMa-7B + SFT and LLaMa-7B + EventRL (Prod-F1). The input text describes the arrest of six individuals by British anti-terror police in Derbyshire and London. While both methods correctly identified the ``ArrestJail'' event, the location (``Derbyshire'', ``London''), and the action (``arrested''), LLaMa-7B + EventRL (Prod-F1) demonstrated a significant improvement by accurately including ``police'' as the agent conducting the arrest. Unlike LLaMa-7B + SFT, which missed the agent's role and the ``crime'' argument, LLaMa-7B + EventRL (Prod-F1)'s result reflects a comprehensive understanding of the event's structure, indicating its superior capability in capturing crucial aspects of events. More case studies can be found in Appendix~\ref{app:analysis_eventrl}.

\section{Conclusion}
In this work, we demonstrated that EventRL, a reinforcement learning approach, significantly enhances the performance of LLMs in event extraction. By focusing on outcome supervision and utilizing specialized reward functions, EventRL effectively addresses the challenges of instruction following and hallucination in event extraction, leading to more accurate and reliable event extraction. The method's success is evident in its superior performance across various model sizes and architectures, particularly in handling novel event types. The importance of selecting appropriate reward functions and the positive impact of code data enhancement on event extraction capabilities have also been highlighted. Furthermore, our findings suggest that while increasing model scale can improve performance, there is a need to balance this with the ability to generalize to avoid overfitting.

\section*{Limitations}
While EventRL is effective in event extraction for LLMs, it faces certain limitations. Firstly, the success of this method heavily depends on the availability of high-quality, well-balanced datasets and meticulous annotation efforts, which can be challenging and resource-intensive. Secondly, as the data volume increases, the training process becomes more time-consuming, necessitating advanced training frameworks and superior hardware capabilities to manage the computational demands efficiently. Lastly, EventRL is specifically designed to address the intricacies of event extraction tasks and does not inherently enhance the general capabilities of large models across a broader spectrum of NLP tasks. This focus on a niche area, while beneficial for its intended purpose, means that the improvements in event understanding and extraction may not translate to a broader enhancement of the models' overall performance in diverse linguistic tasks.

\bibliography{rebiber}

\appendix
\section{Appendix}
\label{sec:appendix}

\subsection{Dataset}
\label{app:dataset}

\paragraph{Dataset Details}

\begin{table}[ht]
\centering
\begin{tabular}{lc}
\hline
\textbf{Dataset} & \textbf{Count} \\
\hline
Train & 285 \\
Dev & 64 \\
Held-in Test & 125 \\
Held-out Test & 380 \\
\hline
\end{tabular}
\caption{Dataset sample counts based on our partitioning of the ACE05 dataset, with a focus on maintaining balance across event types. It is important to note that individual samples may contain multiple event types, leading to shared samples among different event types. This approach ensures a balanced representation of event types, despite variations in sample counts for each type.}
\label{table:dataset_distribution}
\end{table}

\begin{table}[ht]
\centering
\begin{tabular}{lc}
\hline
\textbf{Event Type} & \textbf{Count} \\
\hline
Conflict.Attack & 1244 \\
Movement.Transport & 608 \\
Life.Die & 515 \\
Contact.Meet & 254 \\
Personnel.End-Position & 170 \\
Transaction.Transfer-Money & 167 \\
Personnel.Elect & 153 \\
Life.Injure & 121 \\
Contact.Phone-Write & 112 \\
Transaction.Transfer-Ownership & 109 \\
Personnel.Start-Position & 107 \\
Justice.Charge-Indict & 101 \\
Justice.Trial-Hearing & 97 \\
Justice.Sentence & 93 \\
Justice.Arrest-Jail & 79 \\
\hline
\end{tabular}
\caption{Top 15 event types by sample count in the ACE05 dataset.}
\label{table:top15_event_types}
\end{table}

To provide a detailed and comprehensive evaluation of our model, we conducted a series of experiments using the ACE05 dataset, widely recognized for its variety in event types. This diversity enables us to rigorously test the model's ability to recognize and extract both familiar (seen) and novel (unseen) event types with precision. The ACE05 dataset comprises 33 distinct event types, serving as a robust foundation for our experimental framework.

Our experimental design involved the careful selection of event types for different portions of the dataset: the training set, validation set, and the test sets. Specifically, we chose 7 event types for inclusion in the training and validation sets, as well as the held-in test set. These event types were selected based on their prevalence in the dataset, as indicated by the Table~\ref{table:top15_event_types}. This decision was informed by the observation that the dataset exhibits a significant imbalance in the distribution of event types, with the top 7 event types each having more than 150 samples, while the least represented event type has as few as 2 samples. To ensure a rigorous evaluation of the model’s generalization capabilities, we selected 19 different event types to construct the held-out test set, focusing on unseen event types.

To address the challenge of data imbalance and to maintain a balanced dataset, we adopted a sampling strategy that ensures equitable representation of each event type across different sets. Specifically, we sampled 50 instances for each event type in the training set, 10 for each in the validation set, and 20 for each in both the held-in and held-out test sets. This balanced approach ensures that the model is exposed to and evaluated under a variety of conditions, offering a comprehensive insight into its performance across a wide range of event types.
Table~\ref{table:dataset_distribution} shows the statistics of our ACE05 dataset split. We aimed for balance among event types, although some samples might include several events. This method ensures a fair representation across the dataset.

\paragraph{Event Extraction Guidelines}
Our work is built upon the foundational work of \citet{sainz2023gollie}, which introduced a Python code-based representation for input and output in information extraction tasks. The essence of this work lies in the integration of event type extraction guidelines into the prompt, enhancing zero-shot generalization capabilities.

Our model operates on a schema defined in Python classes, with docstrings providing guidelines and comments outlining representative annotation candidates. This structured format ensures clarity, facilitates parsing, and aligns with modern Large Language Models' (LLMs) pretraining on code datasets. The output, beginning after ``result ='', comprises a list of class instances, yielding a transparent and easily parsable structure when executed in Python.

\begin{figure*}[ht]
    \centering
    \includegraphics[width=1.0\textwidth]{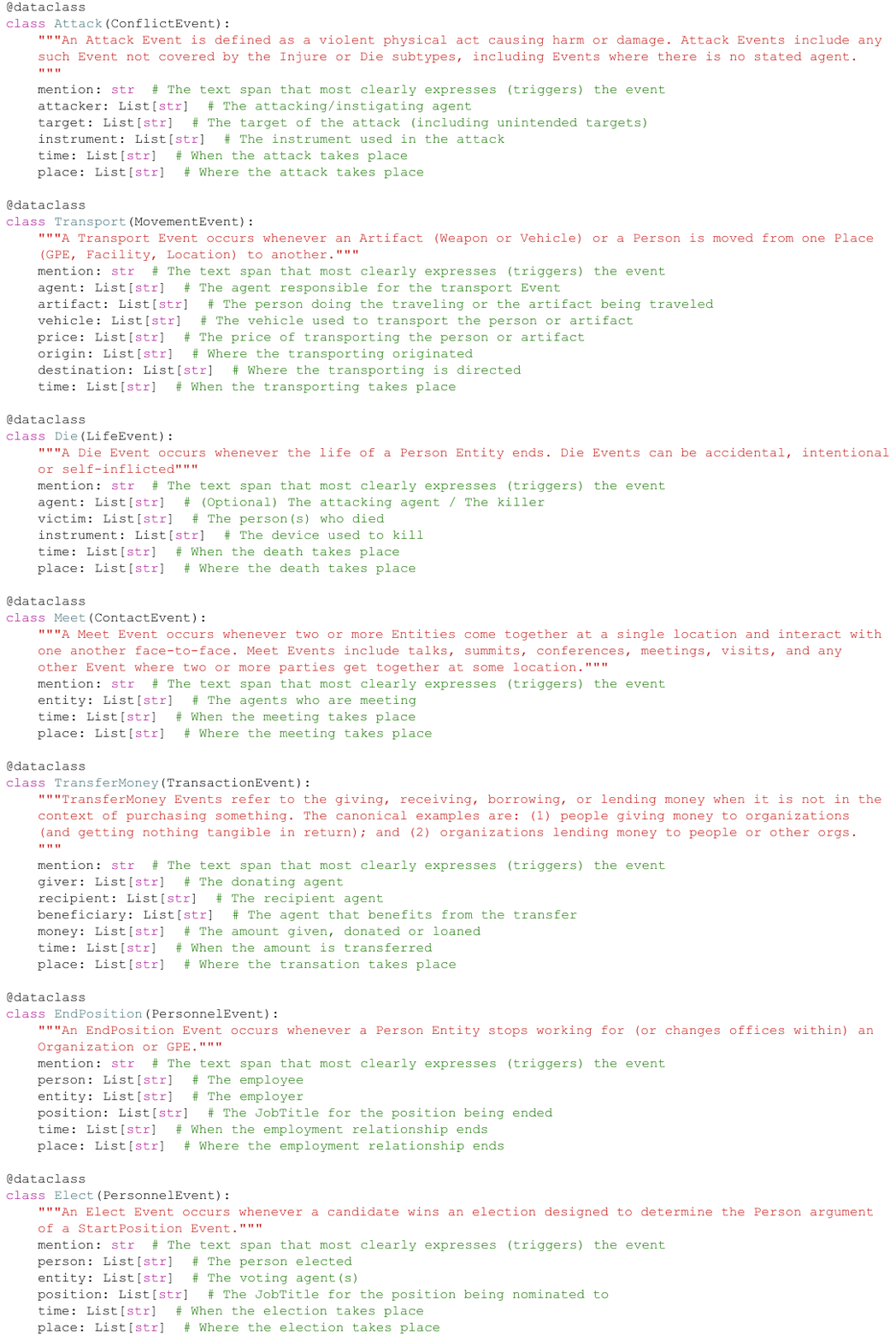}
    \caption{ACE05 Event definitions in Held-in test.}
    \label{fig:definitions}
\end{figure*}

A complete example of our event definitions is displayed in Figure~\ref{fig:definitions} within the Held-in Test, showcasing the guidelines used for extracting event types. We have adapted and refined the data processing code, originally available on the GitHub repository~\footnote{\url{https://github.com/hitz-zentroa/GoLLIE}}, to accommodate our specific needs for event extraction.

\subsection{Implementation Details}
\label{app:implem_details}
Our implementation of EventRL comprises three pivotal components: Few-Shot Prompting (FSP) with GPT-4, Supervised Fine-Tuning (SFT), and the EventRL framework.

\paragraph{GPT-4 and Few-Shot Prompting (FSP)}
For the Few-Shot Prompting experiments, we utilized the GPT-4 API provided by Azure OpenAI Service~\footnote{\url{https://learn.microsoft.com/en-us/azure/ai-services/openai/overview}}, version dated 2023-05-15. Our experiments spanned four settings: 0-shot, 1-shot, 2-shot, and 3-shot. In the 0-shot setup, we did not include any demonstration examples in the instructions. For the 1-shot, 2-shot, and 3-shot setups, we introduced 1, 2, and 3 demonstration examples into the instructions, respectively. We tested across three distinct instruction templates and chose the one that showed the best overall performance for our results' presentation.

\paragraph{Supervised Fine-Tuning (SFT)}
The SFT experiments were conducted on various models: LLaMa-7B/13B and CodeLLaMa-7B/13B/34B. We used the ColossalAI~\footnote{\url{https://github.com/hpcaitech/ColossalAI/tree/main}} framework on two A100 servers. The training setup was as follows: learning rate set to 2e-5, with a minimum learning rate of 2e-6, weight decay at 0.1, micro batch size of 2, global batch size of 64, using bf16 for mixed precision, over 10 training epochs. For the 7B and 13B models, the parallel strategy involved a Tensor Parallel Size of 2 and a Pipeline Parallel Size of 2. The 34B model's strategy was adjusted to a Tensor Parallel Size of 4 and a Pipeline Parallel Size of 2. We selected the best model checkpoint for final use based on its average Trigger-F1 and Argument-F1 scores on the validation set.

\paragraph{EventRL}
Leveraging the SFT groundwork, we proceeded to further train LLaMa-7B/13B and CodeLLaMa-7B/13B models using the EventRL method. The EventRL training was not applied to the 34B model due to computational limits. For the base model selection in EventRL training, we considered the 7B model's lower overfitting risk and the 13B model's higher risk, ultimately choosing the best SFT model for the 7B and the checkpoint from the previous epoch of the best SFT model for the 13B. The EventRL was implemented with the Huggingface transformers~\footnote{\url{https://github.com/huggingface/transformers}} framework, setting the training parameters as follows: a learning rate of 5e-6, a global batch size of 32, a micro batch size of 2, spanning 10 training epochs, using bf16 for mixed precision, and applying advantage clipping at 10. We set the teacher forcing threshold at 70 for the 7B model and 30 for the 13B model. The parameters for generating results with the sampling method were a temperature of 0.5 and a top p of 0.95.


\begin{figure}[ht]
  \centering
  \begin{subfigure}[b]{0.45\textwidth}
    \includegraphics[width=\textwidth]{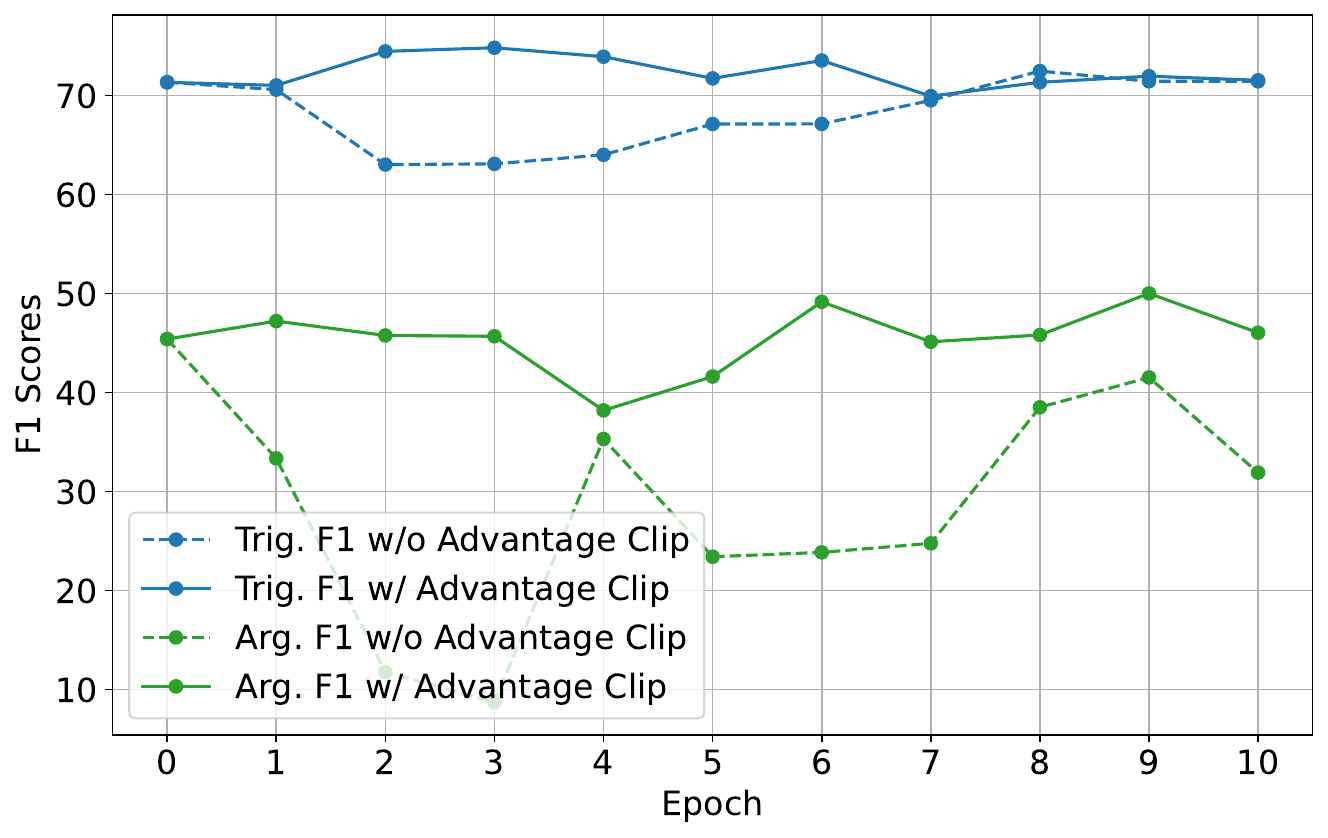}
    \caption{} 
    \label{ablation_aclip_llama7b_role_f1}
  \end{subfigure}
  \begin{subfigure}[b]{0.45\textwidth}
    \includegraphics[width=\textwidth]{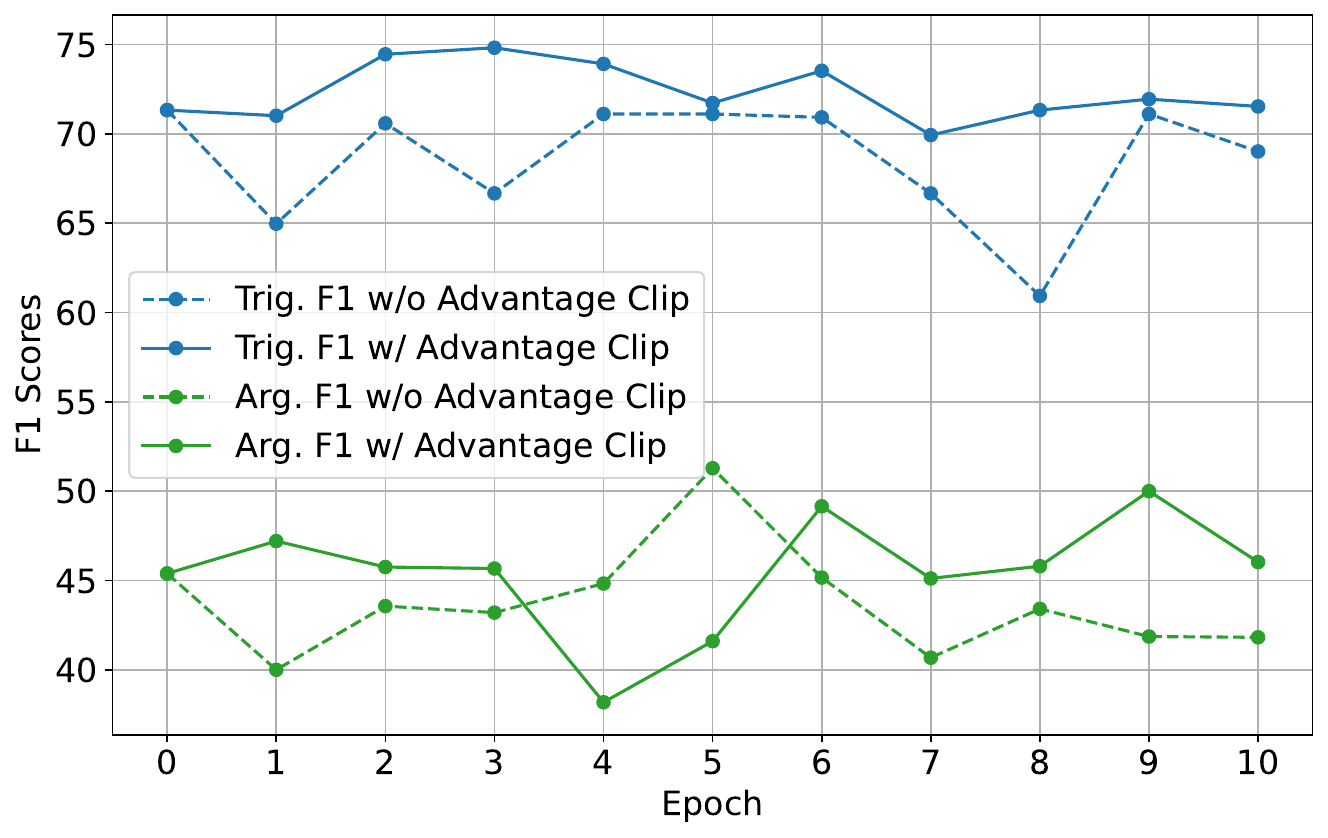}
    \caption{} 
    \label{ablation_aclip_llama7b_product_f1}
  \end{subfigure}
  \caption{Training performance of LLama-7B model on EventRL with and without Advantage Clipping. Figure (a) shows the results for EventRL (Arg-F1) process, and Figure (b) for EventRL (Prod-F1) process. Both graphs compare the Trigger Extraction (Trig. F1) and Argument Extraction (Arg. F1) F1 scores over 10 epochs, illustrating the impact of Advantage Clipping on model stability and learning consistency.}
\end{figure}

  \begin{figure}[ht]
  \begin{subfigure}[b]{0.45\textwidth}
    \includegraphics[width=\textwidth]{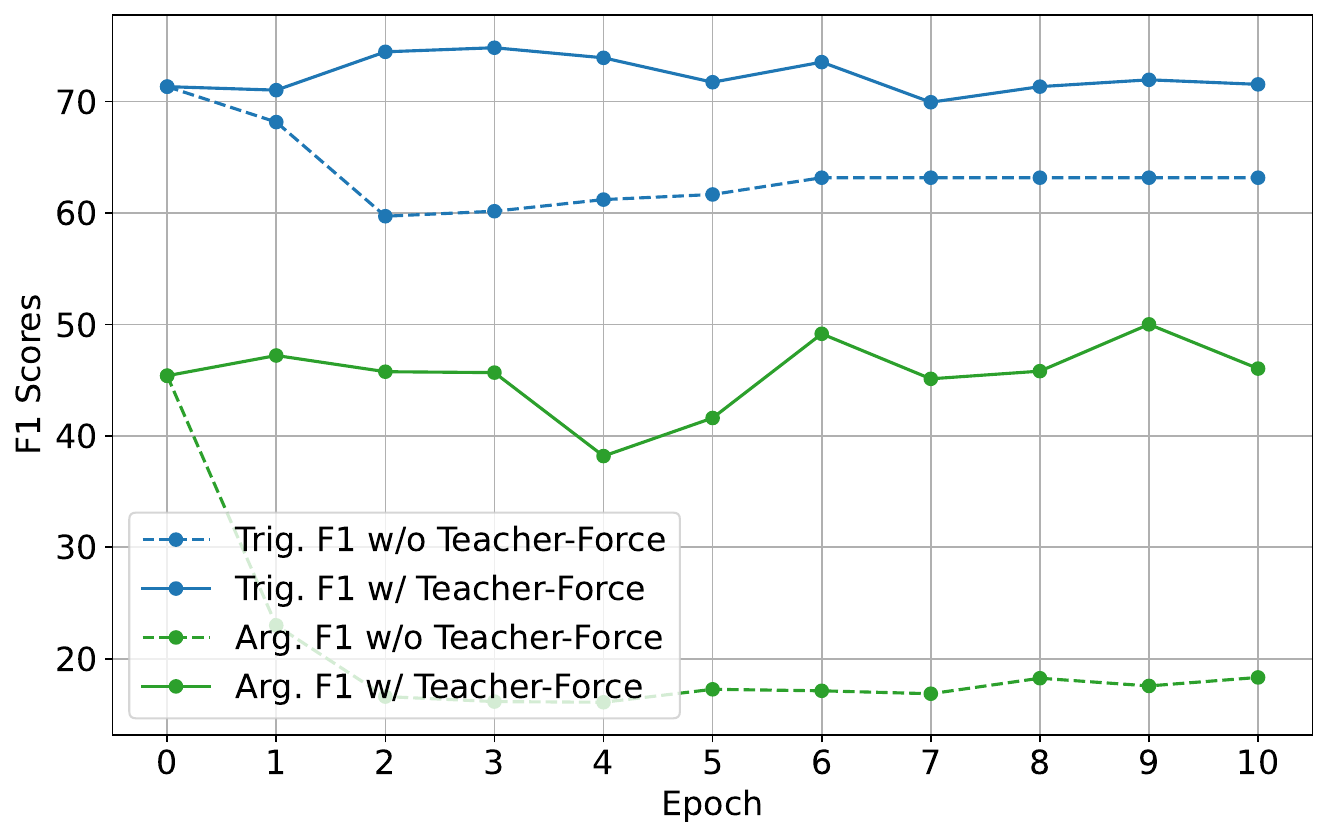}
    \caption{} 
    \label{ablation_teach_llama7b_role_f1}
  \end{subfigure}
  \begin{subfigure}[b]{0.45\textwidth}
    \includegraphics[width=\textwidth]{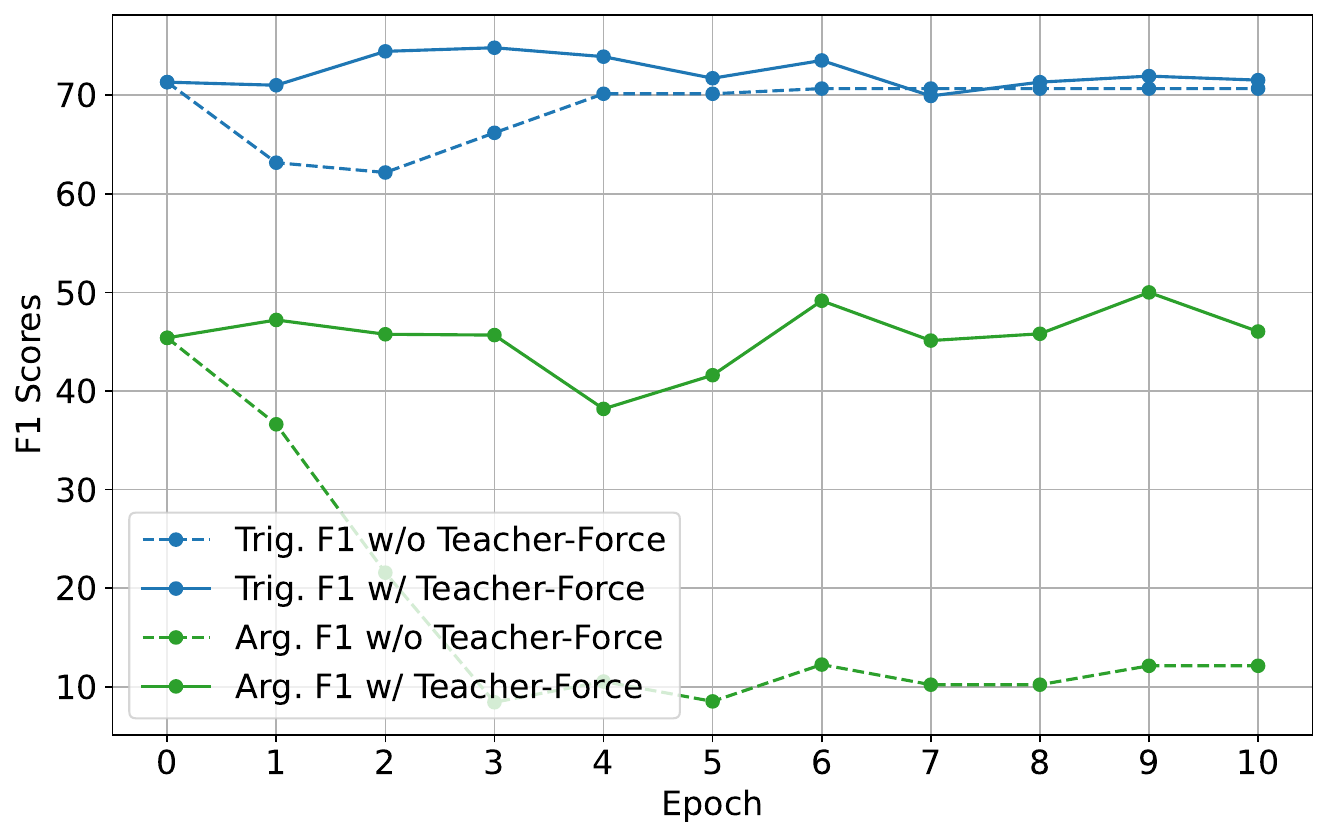}
    \caption{} 
    \label{ablation_teach_llama7b_product_f1}
  \end{subfigure}
  \caption{Training performance of the LLama-7B model employing EventRL strategies over epochs. Figure (a) shows the results using the EventRL (Arg-F1) method, while Figure (b) utilizes the EventRL (Prod-F1) method. Both figures track the Trigger F1 (Trig. F1) and Argument F1 (Arg. F1) metrics across ten epochs. The performance is measured on a validation set} 
\end{figure}

\subsection{More Analysis on EventRL}
\label{app:analysis_eventrl}
\paragraph{Analysis on Teacher Force Threshold}
The Teacher-Force Threshold appears to be a pivotal strategy for stabilizing the training process in EventRL. By examining the Figure~\ref{ablation_teach_llama7b_role_f1} and Figure~\ref{ablation_teach_llama7b_product_f1}, we can infer that the models with Teacher-Force (denoted by ``w/ Teacher-Force'' in the legends) maintain or improve performance consistently across epochs, as opposed to models trained without Teacher-Force (denoted by ``w/o Teacher-Force''), which exhibit more significant fluctuations and generally lower performance scores.

In Figure~\ref{ablation_teach_llama7b_role_f1}, which shows the EventRL (Arg-F1) model's performance, the presence of Teacher-Forcing corresponds with higher and more stable Argument F1 scores. The stability is particularly notable in the later epochs, suggesting that Teacher-Forcing aids the model in retaining knowledge over time, likely mitigating the effects of catastrophic forgetting.
Similarly, in Figure~\ref{ablation_teach_llama7b_product_f1} for the EventRL (Prod-F1) method, we see that the use of Teacher-Forcing correlates with more stable Trigger F1 scores. The Trigger F1 performance without Teacher-Force drops noticeably after the first few epochs, while with Teacher-Force, the performance remains relatively stable.

\begin{figure*}[ht]
    \centering
    \includegraphics[width=1.0\textwidth]{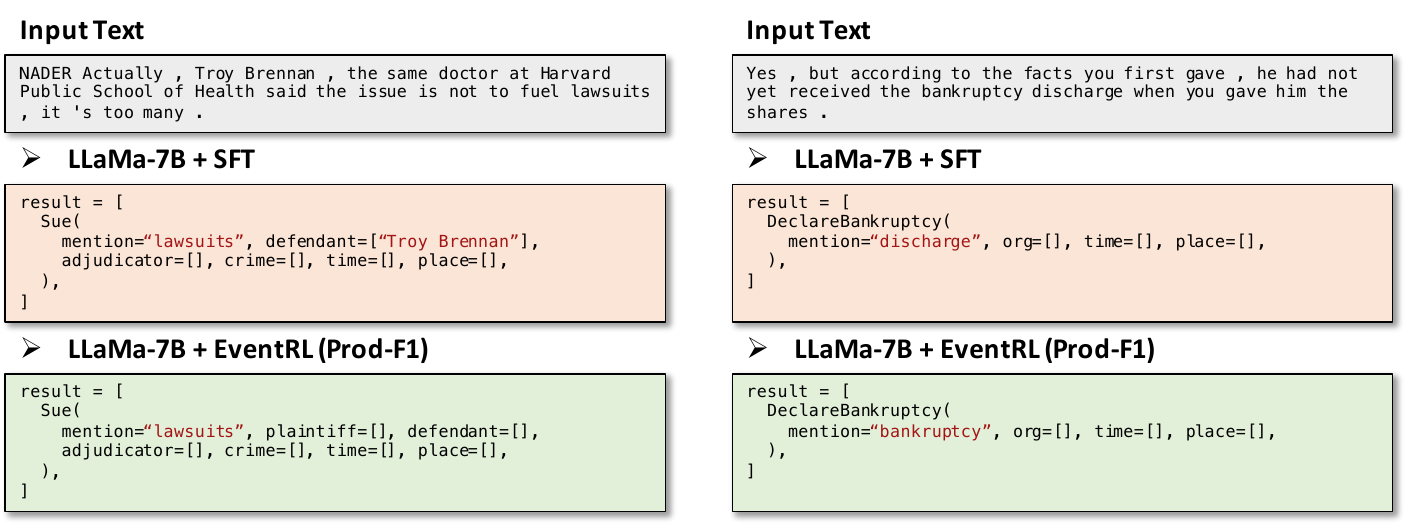}
    \caption{Comparative analysis of event extraction outcomes between LLaMa-7B + SFT and LLaMa-7B + EventRL (Prod-F1) on two distinct cases. Case 1 involves a discussion on lawsuits without explicitly assigning roles, where EventRL accurately avoids misattributing roles unlike SFT. Case 2 focuses on bankruptcy proceedings, with EventRL correctly identifying the primary event mention as ``bankruptcy''. Note that here the results of EventRL (Prod-F1) are totally accurate.}
    \label{fig:case_study2}
\end{figure*}

\begin{table*}[ht]
\centering
\small
\label{tab:gpt4_analysis}
\begin{tabular}{lcccccc}
\toprule
 & \multicolumn{3}{c}{\textbf{Held-in test}} & \multicolumn{3}{c}{\textbf{Held-out test}}\\
\cmidrule(r){2-4} \cmidrule(lr){5-7}
\textbf{Method}& \textbf{Trigger} & \textbf{Argument} & \textbf{AVG} & \textbf{Trigger} & \textbf{Argument} & \textbf{AVG} \\
\midrule
GPT4+FSP (0-shot) + Prompt1 &4.65&22.43&13.54&0.00&5.66&2.83\\
GPT4+FSP (0-shot) + Prompt2 &6.04&22.08&14.06&15.42&17.69&16.56\\
GPT4+FSP (0-shot) + Prompt3 &1.47&21.36&11.41&3.59&18.39&10.99\\
\midrule
GPT4+FSP (1-shot) + Prompt2 + Example1 & 23.02 & 22.82 & 22.92 & 19.32 & 17.83 & 18.58\\
GPT4+FSP (1-shot) + Prompt2 + Example2 &8.63&24.37&16.50&8.87&19.08&13.97\\
GPT4+FSP (1-shot) + Prompt2 + Example3 &15.44&23.33&19.38&12.62&19.50&16.06\\
\bottomrule
\end{tabular}
\caption{Performance comparison of GPT-4 with Few-Shot Prompting (FSP) using different prompts and examples across ``Held-in test'' and ``Held-out test'' settings. The table showcases the Trigger and Argument F1 scores, along with the average (AVG) performance for each method.}
\end{table*}

\paragraph{Analysis on Advantage Clipping}
For both Argument Extraction (Arg. F1) and Trigger Extraction (Trig. F1), the results suggest that Advantage Clipping contributes to more consistent performance across epochs. Specifically, in Figure~\ref{ablation_aclip_llama7b_role_f1}, the use of Advantage Clipping appears to maintain higher F1 scores for Argument Extraction throughout the training epochs, with less variability compared to the setting without Advantage Clipping. The presence of Advantage Clipping seems to prevent drastic drops in performance, which could be indicative of the model retaining previously learned information better, thus mitigating catastrophic forgetting.
Similarly, in Figure~\ref{ablation_aclip_llama7b_product_f1}, for Trigger Extraction, the application of Advantage Clipping demonstrates a more stable and consistently higher performance curve than without it. The variance is visibly reduced, which suggests that Advantage Clipping allows each sample to influence the learning process enough to be remembered, but not so much that it causes significant performance swings.

The most notable observation is the presence of fewer and less severe dips in performance for both Argument and Trigger F1 scores when Advantage Clipping is applied. This smoothing effect implies that Advantage Clipping indeed sets a floor for learning contributions from each sample, ensuring that all training data is utilized effectively, which is particularly crucial for a model that might otherwise focus too heavily on the most recent or the most rewarding examples.

\paragraph{Case Study}
As shown in Figure~\ref{fig:case_study2}, we analyze two instances where the EventRL (Prod-F1) model outperforms the LLaMa-7B + SFT approach.
In the first case, the input text discusses a statement made by Troy Brennan from Harvard Public School of Health regarding lawsuits. The LLaMa-7B + SFT model inaccurately identifies Troy Brennan as a defendant in a lawsuit, reflecting a misunderstanding of the context. In contrast, the LLaMa-7B + EventRL (Prod-F1) model provides a more accurate representation by not assigning specific roles to the entities involved in the ``Sue'' event. This output aligns better with the input text, which does not explicitly define plaintiffs or defendants but rather discusses the broader issue of lawsuits.

The second input text refers to a situation involving bankruptcy proceedings. The LLaMa-7B + SFT model incorrectly identifies ``discharge'' as the event mention, which misrepresents the text's focus. The term "discharge" in the context of bankruptcy refers to the legal process of releasing a debtor from certain obligations, but the key event is the declaration of bankruptcy itself. The LLaMa-7B + EventRL (Prod-F1) model correctly identifies ``bankruptcy'' as the event mention, providing a more accurate and relevant extraction.

\subsection{More Analysis on GPT4+FSP}

\begin{figure*}[ht]
    \centering
    \includegraphics[width=1.0\textwidth]{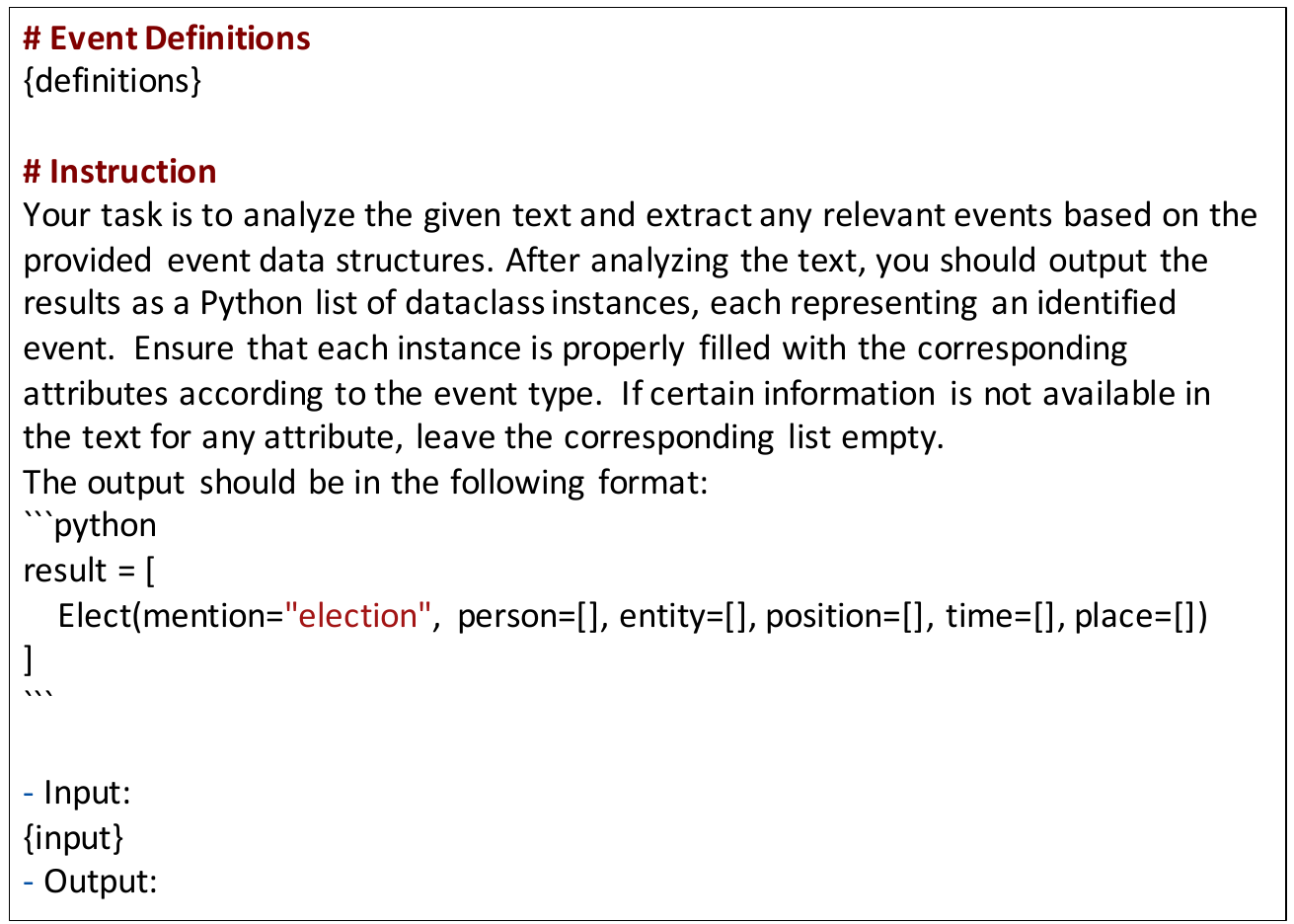}
    \caption{Template 1 illustrates a structured approach to event extraction, combining Python dataclass definitions with clear natural language instructions. It emphasizes a precise output format, guiding users on how to represent extracted events as a list of dataclass instances.}
    \label{fig:gpt4_temp1}
\end{figure*}

\begin{figure*}[ht]
    \centering
    \includegraphics[width=1.0\textwidth]{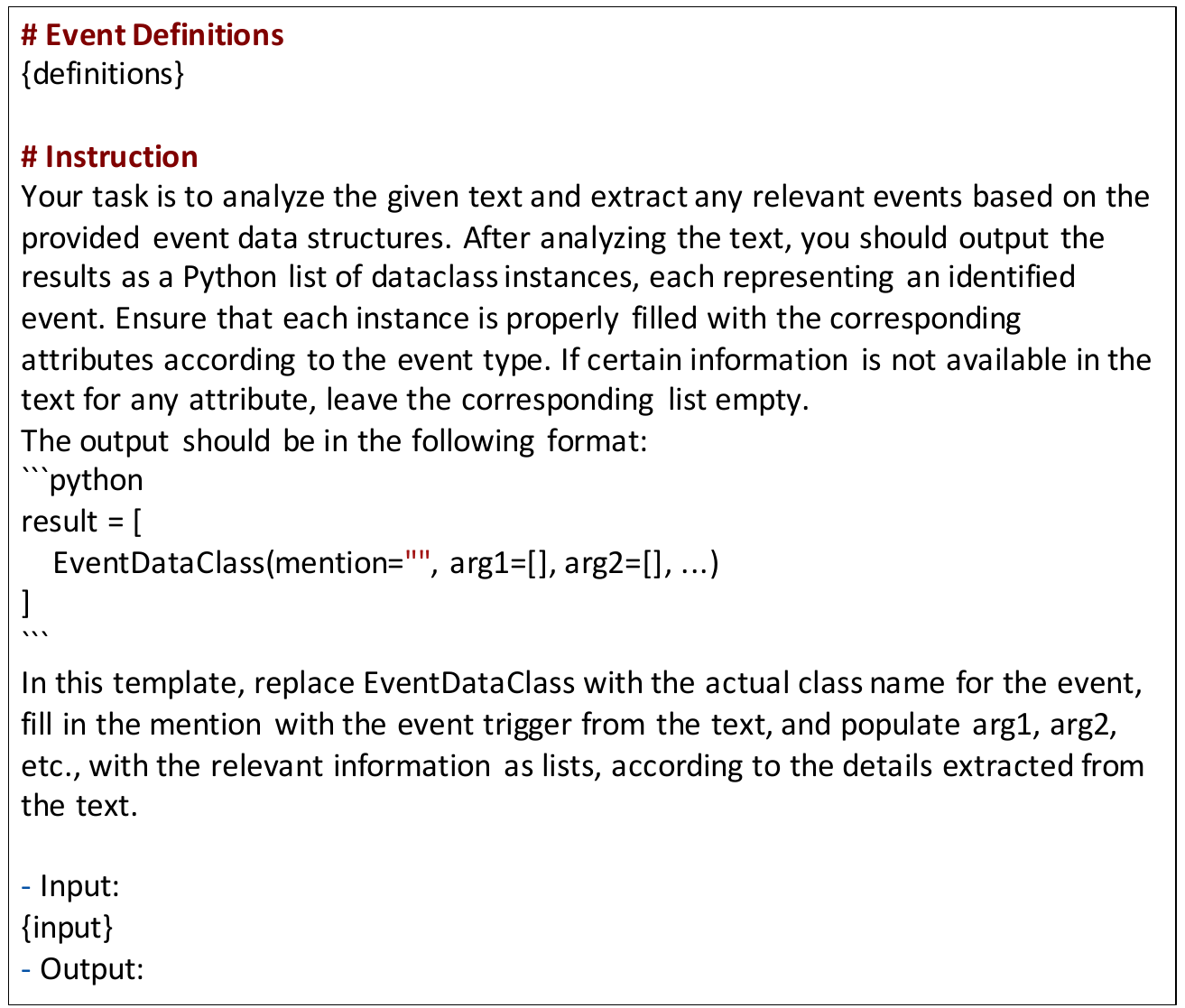}
    \caption{Template 2 presents a hybrid format that merges programming structure with user-friendly instructions for event extraction. It details how to fill out dataclass instances based on text analysis, with a focus on maintaining a clear and standardized output format.}
    \label{fig:gpt4_temp2}
\end{figure*}

\begin{figure*}[ht]
    \centering
    \includegraphics[width=1.0\textwidth]{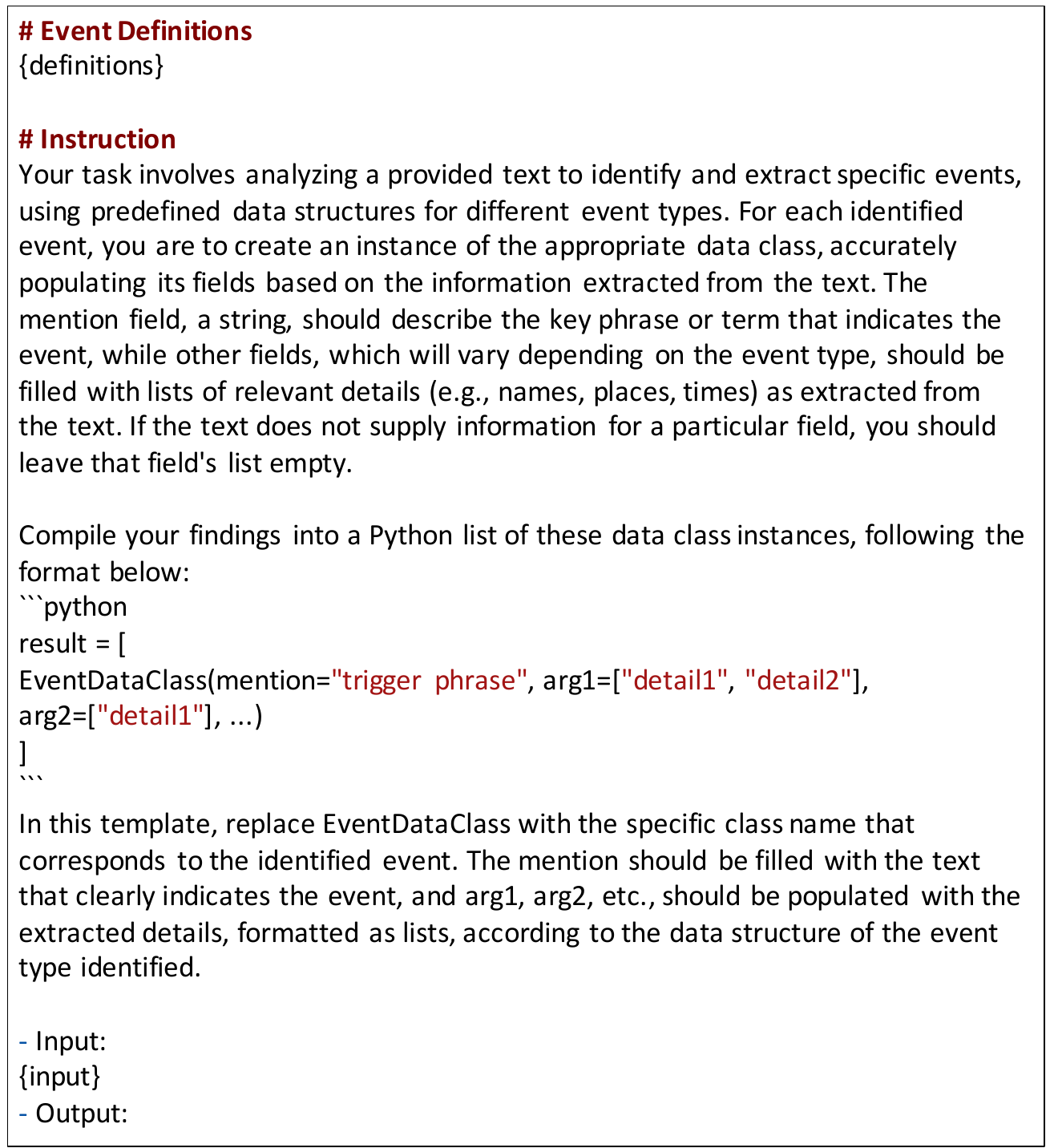}
    \caption{Template 3 offers a comprehensive guide for identifying and extracting events from text, using predefined data structures. It specifies how to populate data class fields with extracted details, aiming to enhance clarity and accuracy in representing events through a list of dataclass instances.}
    \label{fig:gpt4_temp3}
\end{figure*}

\paragraph{Analyzing Various Prompts for GPT4+FSP (0-shot)}
The performance of GPT-4 with Few-Shot Prompting (0-shot) varies significantly across different prompt templates, illustrating the impact of instructional design on the model's ability to extract events. Prompt 2~(See Figure~\ref{fig:gpt4_temp2}), which provides a clear and structured task description along with an explicit output format, yields the highest overall performance, especially noticeable in the ``Held-out test'' section with a notable average score of 16.56. This suggests that the clarity and specificity of the instructions can significantly enhance the model's understanding and execution of the task. In contrast, Prompts 1~(See Figure~\ref{fig:gpt4_temp1}) and 3~(See Figure~\ref{fig:gpt4_temp3}), despite offering detailed instructions, do not match the effectiveness of Prompt 2, potentially due to differences in how the task and output format are communicated.

\paragraph{Analyzing Different Examples for GPT4+FSP (1-shot)}
Introducing examples in the Few-Shot Prompting (1-shot) setup with GPT-4 shows a nuanced effect on performance, underscoring the influence of example selection. The inclusion of Example 1 with Prompt 2 significantly boosts performance across both ``Held-in test'' and ``Held-out test'', achieving an average score of 18.58 in the latter. This improvement indicates that the right example can enhance the model's understanding of the task, leading to better event extraction outcomes. However, the impact of different examples varies, with Example 2 and Example 3 leading to mixed results. 

\subsection{Discussion}
One key observation from our study is the difference in performance between large and small models. Large models tend to perform better because they have more capacity to understand and process complex information. This means they can better identify the events in texts and figure out the relationships between different parts of the information.
However, not everyone can use large models because they need a lot of computing power and resources. This is where our EventRL framework comes into play. EventRL is designed to help smaller models perform better on complex tasks. It does this by focusing on the outcomes of the task and using reinforcement learning to guide the model towards better performance.

With EventRL, even smaller models can improve their ability to extract events from texts. The approach helps these models pay closer attention to the final goal of the task and learn from each attempt. This way, they get better over time at understanding what's important in the text and how to accurately identify events and their details.


\end{document}